\documentclass[letterpaper, 10 pt, conference]{ieeeconf}  % Comment this line out if you need a4paper

\IEEEoverridecommandlockouts                              % This command is only needed if 
\usepackage{cite}
\usepackage{amsmath,amssymb,amsfonts}
\usepackage{algorithmic}
\usepackage{graphicx}
\usepackage{textcomp}
\usepackage{xcolor}
\usepackage{threeparttable}
\usepackage{multirow}
\usepackage{multirow}
\usepackage{hyperref}

\hypersetup{%
 hidelinks,
 colorlinks=false}

\title{\LARGE \bf
Underwater Soft Fin Flapping Motion \\ with Deep Neural Network Based Surrogate Model
}

\author{Yuya Hamamatsu$^{1}$, Pavlo Kupyn$^{1, 2}$, Roza Gkliva$^{1}$, Asko Ristolainen$^{1}$, Maarja Kruusmaa$^{1}$
\thanks{$^{1}$The authors are with the Department of Computer Systems, Tallinn University of Technology, Tallinn, Estonia
        {\tt\small (Yuya.Hamamatsu, Pavlo.Kupyn, Roza.Gkliva, Asko.Ristolainen,
        Maarja.Kruusmaa)@taltech.ee}, $^{2}$ Siauliai Academy, Vilnius University}
}

\begin{document}

\maketitle
\thispagestyle{empty}
\pagestyle{empty}

% As a general rule, do not put math, special symbols or citations
% in the abstract or keywords.
\begin{abstract}
This study presents a novel framework for precise force control of fin-actuated underwater robots by integrating a deep neural network (DNN)-based surrogate model with reinforcement learning (RL). To address the complex interactions with the underwater environment and the high experimental costs, a DNN surrogate model acts as a simulator for enabling efficient training for the RL agent. Additionally, grid-switching control is applied to select optimized models for specific force reference ranges, improving control accuracy and stability. Experimental results show that the RL agent, trained in the surrogate simulation, generates complex thrust motions and achieves precise control of a real soft fin actuator. This approach provides an efficient control solution for fin-actuated robots in challenging underwater environments.
\end{abstract}

\section{Introduction}
The designs of underwater robots vary depending on their purpose, and the types of actuators used demonstrate a wide range of designs and characteristics \cite{bogue2015underwater}. Soft-fin-actuated robots have advantages in safety and quietness, making them ideal for sensitive environments such as ocean exploration and wildlife monitoring. However, controlling these fin actuators introduces significant challenges because of their highly complex dynamics \cite{remmas2021inverse}. These complexities also increase from the inherent flexibility of soft materials, which respond to the hydrodynamic resistance force. Fins made of such materials interact dynamically, making their behavior difficult to model with traditional physics-based approaches. Moreover, previous studies of soft actuator motion generation imply advanced knowledge of kinematics and physics modeling \cite{GEORGIADES200939}. Therefore, it is challenging to design the motion trajectories of the soft fins that generate the forces required by the robot, which requires advanced mathematical modeling.

\begin{figure}[t]
\includegraphics[width=0.95\linewidth]{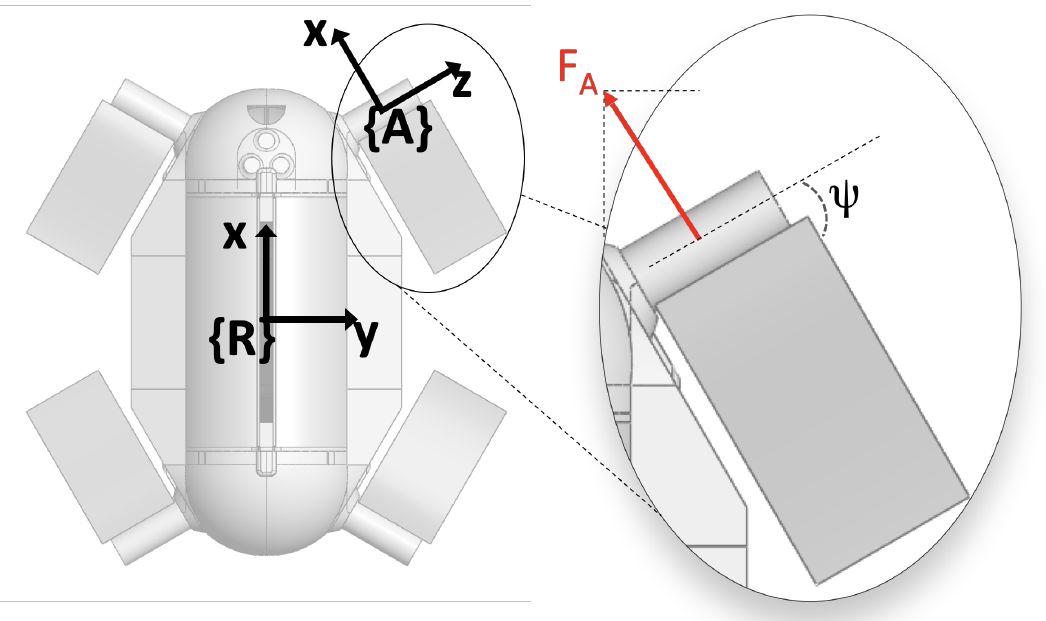}
\caption{An example of the fin actuated robot with four individually actuated soft flippers. $F_A$ is the overall vector of the generated force in the direction of the fin's $x$-axis.}
\label{fig:ucat}
\end{figure}

This study addresses these challenges of soft fin-actuated robots by developing a force model based on a Deep Neural Network (DNN) \cite{kutz2017deep}. The DNN approximates the complex relationship between motor movement and the resulting fin-generated forces, using real-world data for training to develop an accurate surrogate model-based simulator (Real2Sim) \cite{thakur2022deep}, \cite{sun2020surrogate}. Our surrogate simulator is generally usable for multiple different applications consisting of DNN models trained by randomly generated commands. Furthermore, we propose a Reinforcement Learning (RL) agent as a fin motion controller to generate the reference force within this surrogate simulation environment. The use of the simulator can eliminate the need to reconfigure the experiment for each required application, but we can train the motion controller in a more realistic environment. Taking advantage of this ease of training, a grid-switching RL is proposed for the RL design. It trains one RL agent for each reference force, switching models on demand. This approach requires a large amount of training, but we can train them easily and quickly in a realistic environment. To integrate an RL-based approach and a surrogate simulator, the optimal action can be selected to generate force. The real-world transferability of this trained RL controller is implemented in the actual fin actuator and is evaluated (Sim2Real) using the fin actuated robot shown in Fig. \ref{fig:ucat}. %More about robot design in\cite{ucat2014design}.

\begin{figure*}[t]
\includegraphics[clip, width=18cm]{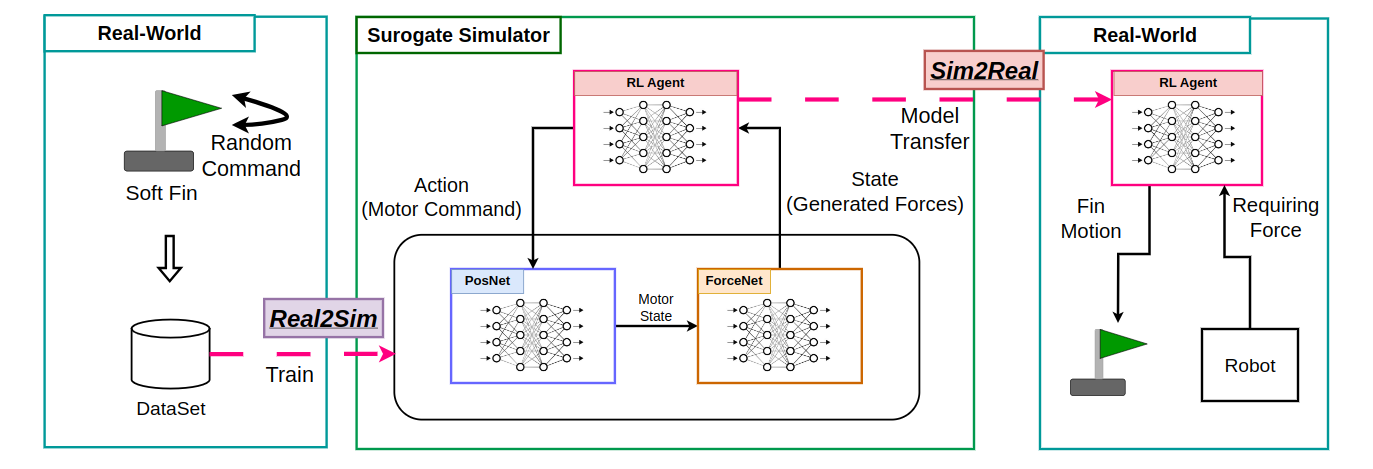}
\caption{Method overview. The surrogate simulation model is trained with a dataset that includes fin motion and force measured by a force sensor. The reinforcement learning agent is trained inside this simulator. The trained model is directly applied to the real fin actuator. (Real2Sim2Real)}
\label{overview}
\end{figure*}

\section{Related Works}
The modeling of soft actuators, particularly in underwater environments, presents significant challenges due to the complexity of soft materials \cite{muhammad2014non}. Traditional physics-based or analytical methods often struggle to accurately capture the complex fluid-structure interactions between soft fins and the surrounding water \cite{singh2019dynamic}. This complexity is further compounded by the complex dynamics that arises from the interaction of flexible fins with water. Early studies on fin actuators based on specially designed oscillate motion with a mathematical model \cite{gkliva2018development}. However, recent advances in data-driven modeling, particularly through machine learning techniques, have yielded more accurate and flexible dynamic models \cite{lee2023data}, \cite{9652036}. These models, such as neural networks, can be trained on empirical data, allowing them to learn the complex, nonlinear behavior of soft actuators without the need for explicit mathematical models, significantly improving adaptability to diverse environmental conditions.

Reinforcement Learning (RL) has also emerged as an effective approach to develop controllers in systems where traditional control methods struggle, especially in complex environments \cite{tong2023survey}, \cite{6315769}. RL-based controllers allow systems to learn optimal policies through trial and error, guided by environment feedback \cite{singh2022reinforcement}. However, in the context of soft robotics, particularly soft underwater fin actuators, the highly complex dynamics pose significant challenges to the direct application of RL. To address these challenges, the integration of a surrogate models within RL frameworks has gained increasing attention, particularly in large-scale control problems such as plant energy management \cite{pinto2021data}, \cite{wang2021surrogate}. Surrogate models, often constructed using neural networks, approximate the behavior of complex systems by predicting outcomes based on precollected data \cite{hou2022dimensionality}. 

In this paper, we apply both surrogate modeling and RL techniques to develop a robust controller for soft-fin-actuated underwater robots. By integrating these approaches, we achieve optimal force generating motion for soft fin actuators.

\section{Method}

In this section, we describe the method for the motion controller to generate forces. The key concept diagram is shown in Fig. \ref{overview}. The Real2Sim2Real configuration is where the simulator is developed using real-world data and the motion controllers trained in this simulator are reapplied to the actuator in the real world.

\begin{figure}[bt]
\includegraphics[width=\linewidth]{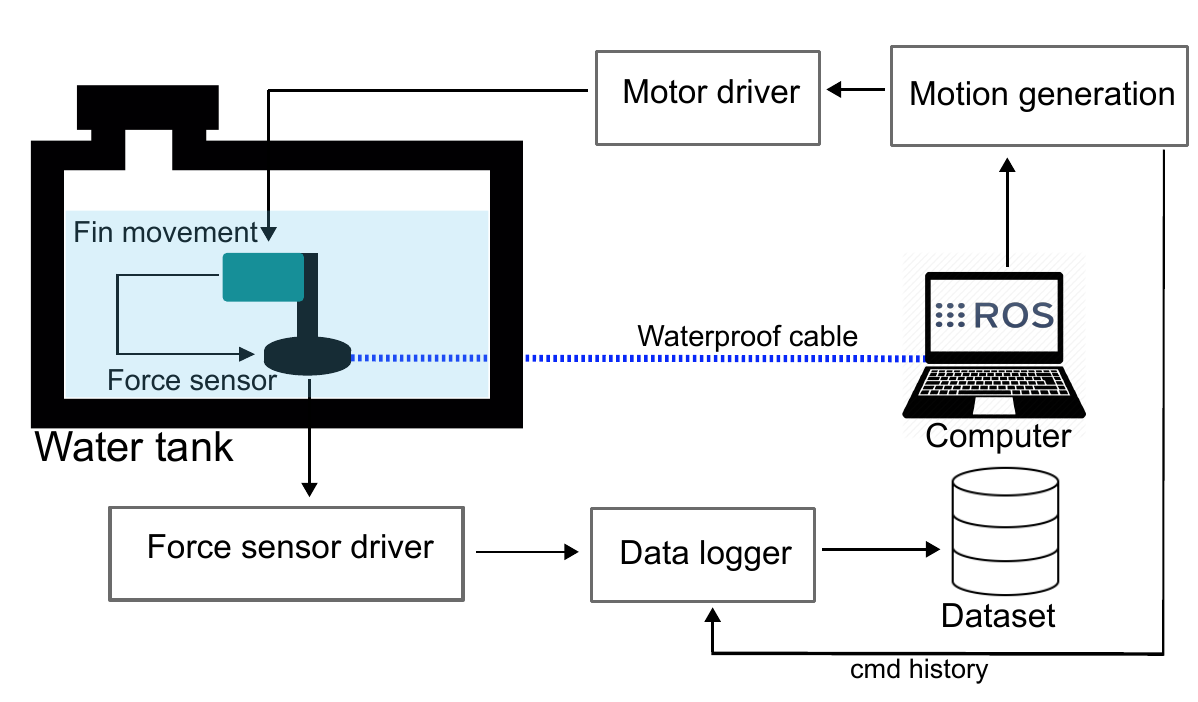}
\caption{Experimental setup: ROS2 computer sends motor commands to the fin motor driver, that generates movement and produces forces that are measured by the force sensor.}
\label{fig:experimental_setup}
\end{figure} 

\begin{figure}[t]
    \centering
    \includegraphics[width=\linewidth]{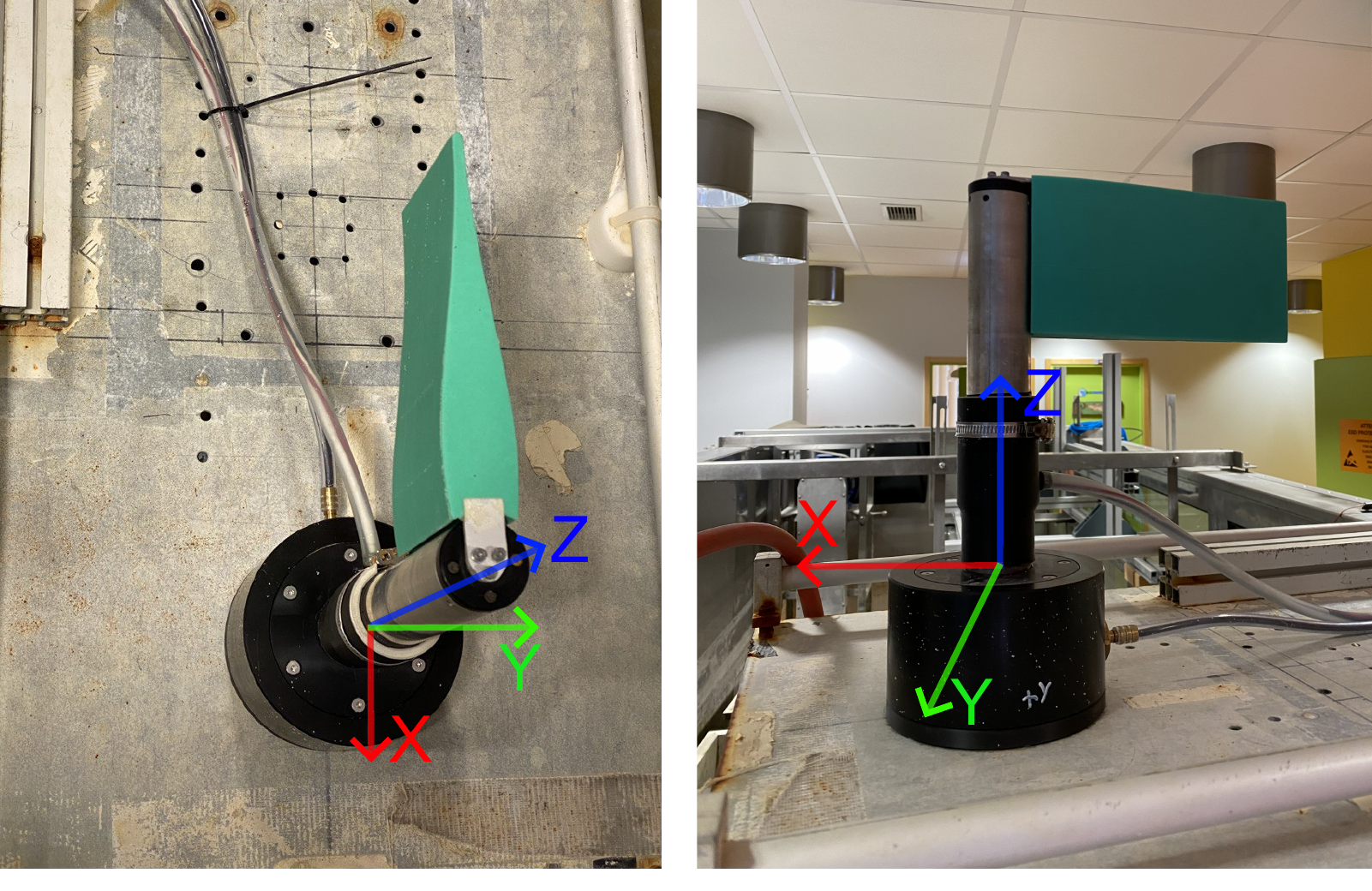}
    \caption{Fin actuator and force sensor and axes from top and side view.}
    \label{fig:fin_axes}
\end{figure}

\subsection{Data collection}
Experiments are conducted in a water tank with still water (Fig.~\ref{fig:experimental_setup}). 
The actuator consists of a soft silicone fin (Zhermack Elite Double 22), cast around a short aluminium bracket. The bracket is fastened to a servo motor shaft (Maxon EC-max 30 brushless EC motor with Hall encoder) and rotates on command (Fig.~\ref{fig:fin_axes}). The motor's encoder is used to obtain the angular position and velocity of the fin. Using a target angle and a target angular velocity as input, the motor driver (EPOS2 Module 36/2 digital positioning controller) moves the motor at the angular velocity specified for the target angle.
% A Rohde \& Schwarz HMP4040 power supply provides constant 28V to the motor driver.

To obtain dynamics data, the motor was mounted on a waterproofed force-torque sensor (Axia80-M20, 	
ATI Industrial Automation), which was then mounted on a metal plate. Both motor and the force sensor are connected to a computer running ROS2 in Ubuntu 22.04. To develop a more generally usable simulator, the motor takes as input a random target position in the range ($-\pi/2$, $\pi/2$) and moves with a random angular velocity in the range (1, $\pi$). During the experiment, position feedback was collected through transmitted commands and encoder information, and force was collected through force sensors. All data is logged back to the computer.
Each data log consists of 20,000 sensor readings at 100 Hz. Overall 100 data logs have been collected. Therefore in total the dataset contains 2,000,000 training examples. Three additional datasets were collected as a test dataset for the evaluation of the surrogate model described in the next section.

\subsection{Surrogate soft-fin dynamics simulator}

\begin{figure}[b]
\includegraphics[width=\linewidth ]{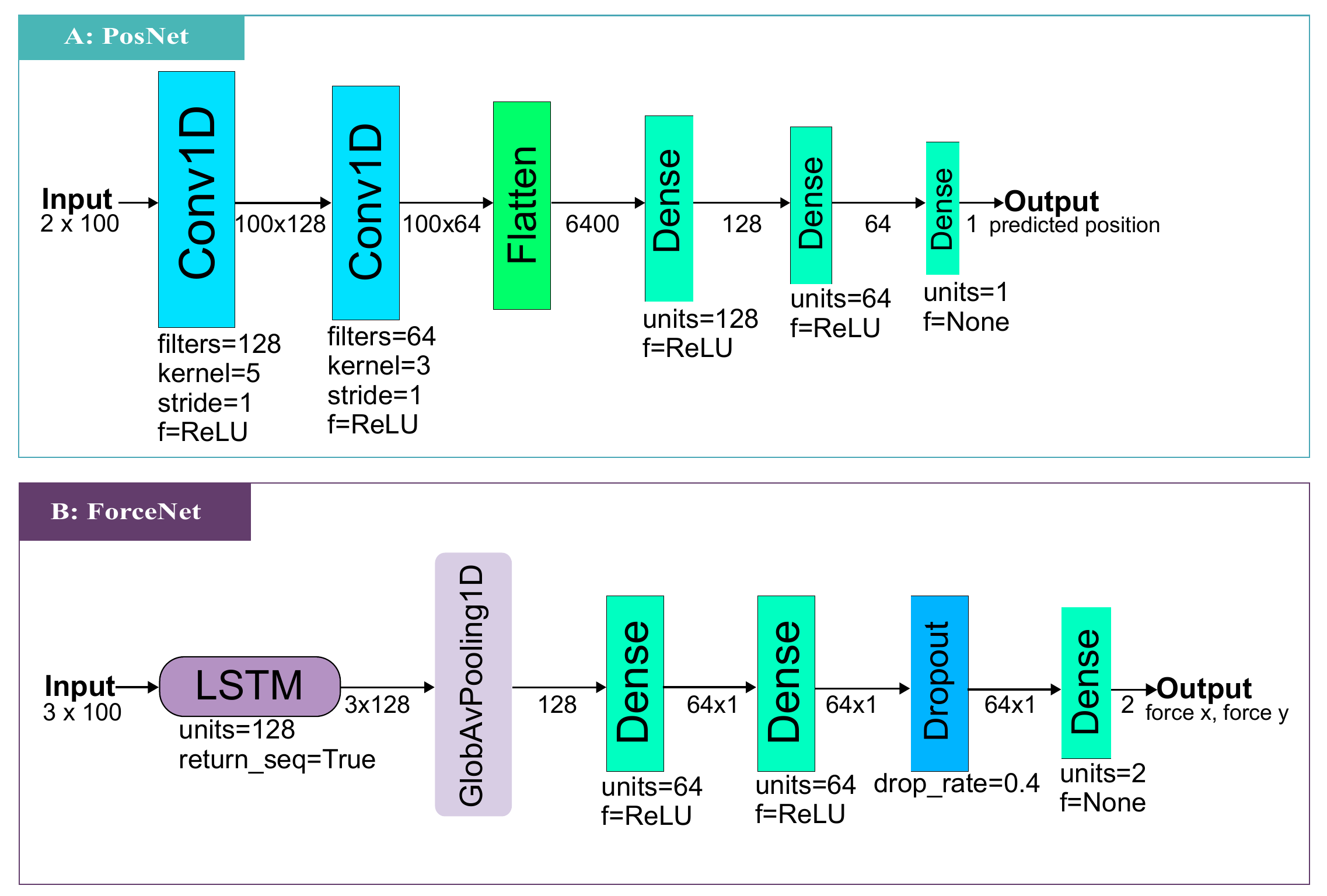}
\caption{Full architecture of surrogate model: (A) PosNet: two convolutional layers followed by three linear layers. (B) ForceNet: LSTM layer followed by two linear layers, dropout and final ouput linear layer.}
\label{fig:surrogate_architecture}
\end{figure}

This section outlines the architecture of a neural network model designed to predict actual force output. Figure \ref{fig:surrogate_architecture} demonstrates the complete architecture pipeline for surrogate model. The relationships between the generated forces and the fin motion command are learned from real datasets, allowing the surrogate model to simulate the behavior of the system without the need for continuous real-world testing which is especially challenging for underwater environments. 

The forward model is composed of two independent neural networks, each responsible for different aspects of motor characteristic and force generated by fin motion. The first network, PosNet, takes the motor commands as input and predicts the actual motion of the motor. This is a critical step as motor behavior often contains inherent uncertainties and variations that make direct prediction difficult. By isolating this aspect, PosNet captures the fin's actual movements based on the input commands. The second network, ForceNet, takes the motor motion data as input and predicts the two-dimensional force (x, y) generated by the fin. This force prediction is nonlinear and varies according to the fin's motion, making ForceNet essential in bridging the gap between motor behavior and the resulting forces. By handling these two tasks separately, the forward model efficiently addresses both the uncertainty in the motor's characteristics and the complexity of the force generation process.

The architecture of PosNet is the one-dimensional convolutional neural network (1D CNN) \cite{cnn2014intro, cnn2012speechrecognition}. The 1D CNN is particularly well-suited for time-series data such as motor commands because of its ability to capture local temporal dependencies while maintaining computational efficiency. Among the advantages of 1D CNN based model architecture are the efficient feature extraction by using convolutional filters, parallel processing of input sequences, and increased training speed. By leveraging these strengths, PosNet efficiently models the motor’s kinematic response to given commands, ensuring that the predictions capture the necessary details despite the presence of motor uncertainty.

ForceNet utilizes the Long Short-Term Memory (LSTM) network, which is well-suited for processing time-series data. The ability of LSTM to maintain information in long sequences ensures that ForceNet can accurately model the temporal dependencies inherent in force output. The retention of time-series information by LSTM is suitable for capturing the effects of soft fin deformation caused by fluid drag at the previous moment, and changes in drag and lift forces due to the flow generated.
%This data-driven approach allows the forward model to generalize well, even in the face of complex, real-world interactions where continuous experimentation would be infeasible.

\subsection{Reinforcement learning based motion acquisition}
In this section, we describe how the surrogate model described in the previous section is integrated into the RL framework and its architecture.
\subsubsection{Surrogate Model Integration for RL}
The RL agent generates an action based on its policy in each timestep. In this study, this action is updated at 3.0 Hz. Instead of directly training this policy using the actual fin and sensors, we use the DNN based surrogate simulator as training environment. The action consists of a motor command first passed to a PosNet. It simulates motor movements against the command. The simulated fin movements by PosNet go into ForceNet. ForceNet estimates the force generated by the fin based on the positional data provided by PosNet. The simulation of the dynamics by the surrogate model is carried out at 100 Hz based on the data acquisition environment. The combination of PosNet and ForceNet allow the RL agent to train the policy and interactions with the environment without physically executing those actions. By embedding this surrogate model within the RL loop, we effectively reduce the need for experiments with the real system during the training phase, significantly accelerating the learning process while maintaining high accuracy in force predictions.

\begin{figure}
\includegraphics[width=0.95\linewidth]{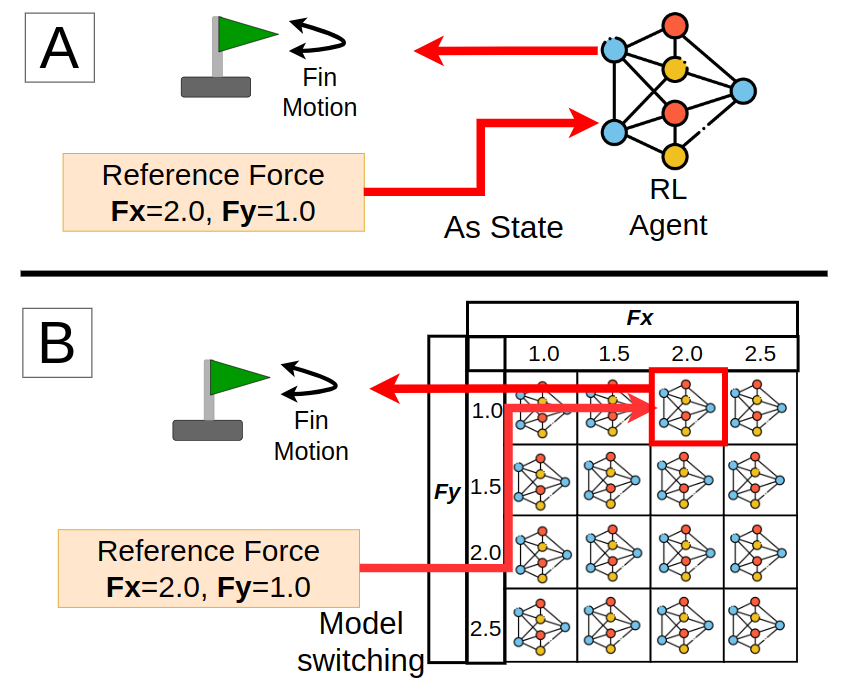}
\caption{Motion controller structure: (A) General single RL agent design. It takes a force reference as the state, and the RL agent generates the motion depending on the state. (B) Grid-Switching design. Switches the model trained separately for each reference force. }
\label{fig:RL}
\end{figure}

\subsubsection{RL algorithm}
The action space in the timestep $t$ ($\mathcal{A}_t$) consists of a target angle ($\theta_t$) and a target angular velocity ($\omega_t$) that are passed as input to PosNet. The state space on timestep $t$ is described as:
\begin{equation}
\mathcal{S}_t= [\theta, F^r_x, F^r_y, \mathcal{A}_t, \mathcal{A}_{t-1},...\mathcal{A}_{t-k}] \text{ ,}
\end{equation}
\noindent where $\theta$ is the last fin angle measured by PosNet (on simulator) or the motor's encoder (on real-world implementation). $F^r_x$ and $F^r_y$ are the reference forces in each direction. $k$ is the parameter that defines how many past actions we keep as the state. The agent then uses its learned policy to choose the next action, which is the motor command that controls the fin movement. 

The reward function in each step $r_{step}$ is based on the error between the predicted force (by ForceNet) and the reference force to generate the optimal force by the movement of the fin. In addition to these errors, rapid changes of the generated force possibly cause unsuitability of the robot attitude. Therefore, we calculate the Sobolev norm to evaluate the stability of force generation \cite{adams2003sobolev}. The Sobolev norm is a mathematical metric for evaluating the smoothness of a function. Specifically, it allows an expression of how close a function is to a target value while also capturing the function's smoothness, meaning its lack of abrupt changes. In the Sobolev space $W^{k,p}(\Omega)$, the first-order Sobolev norm $W^{1,2}$ is defined as follows:

\begin{equation}
\|f \|_{W^{1,2}} = \left( \int_{\Omega} |f(x)|^2 \, dx + \int_{\Omega} |f'(x)|^2 \, dx \right)^{1/2}
\end{equation}

Here, the first term $|f(x)|^2$ represents the norm of the function itself, and the second term $|f'(x)|^2$ represents the norm of its derivative, which captures the smoothness of the function. In implementation, to apply the Sobolev norm to discrete data, smoothness can be approximated using the differences between consecutive values. Specifically, given a set of force data $F^i$ at $i$, a discretized form of the first-order Sobolev norm can be represented as:

\begin{equation}
\| F \|_{W^{1,2}} \approx \left( \sum_{i=0}^{n-1} |F^i - F^r|^2 + \sum_{i=0}^{n-2} |F^{i+1} - F^i|^2 \right)^{1/2}
\end{equation}

Here, the first term is modified to utilise the difference ($F_e$) between the average force and the reference force within the evaluation window (window size 200). This allows the evaluation to achieve both a global target force and a stable output. The reward function is as follows.

\begin{equation}
    r_{\text{step}} = -\sum_{d \in \{x, y\}} w_d \left( |F_e^d| + \lambda_d \sqrt{\sum_{i=0}^{n-2} |F_{i+1}^d - F_i^d|^2} \right)
\end{equation}

% \begin{equation}
% \begin{aligned}
% r^{\text{step}}_x &= - |F^e_x| - \lambda_x \left( \sum_{i=0}^{n-2} |F^{i+1}_x - F^{i}_x|^2 \right)^{1/2} \\
% r^{\text{step}}_y &= -|F^e_y| - \lambda_y \left( \sum_{i=0}^{n-2} |F^{i+1}_y- F^{i}_y|^2 \right)^{1/2}
% \end{aligned}
% \end{equation}

% \begin{equation}
% r_{step}= w_1 r^{\text{step}}_x + w_2 r^{\text{step}}_y
% \end{equation}

This design optimizes both the closeness of the agent's output to the target value and the smoothness of the force outputs. The smoothness regularization term, or Sobolev term, ensures that the agent's outputs do not vary excessively, leading to more stable force control.

As a learning algorithm, we apply LSTM-PPO. The Proximal Policy Optimization (PPO) algorithm is chosen to optimize the RL policy. PPO is good at making stable improvements to the policy without making large updates that could harm performance \cite{schulman2017proximal}. LSTM is used because it can remember past information. For fin motion controller, the fin's dynamics constantly change, and LSTM helps keep track of these changes so the RL agent can make decisions based on previous information. By combining LSTM and PPO, the system can learn to control forces in dynamic environments such as underwater robots.

\begin{table}[t]
\caption{Models Inference}
\label{tab:inf}
\begin{minipage}{\linewidth}
\centering
\begin{tabular}{c c c c c c}
\hline \\
   {\textbf{Model}} & {Window} & {Parameters} & RMSE & MAE & DTW \\[5pt]
\hline \\
  \textbf{PosNet 1D CNN} & 100 & 113601 & 0.03 & 0.002 & 0.002 \\[5pt]

  \textbf{ForceNet LSTM} & 100 & 129794 & 0.2 & 0.04 & 0.03 \\
  \hline \\
\end{tabular}
\end{minipage}
\end{table}

\begin{table*}[t]
\caption{Experiments summary}
\label{tab:eval}
\begin{minipage}{\linewidth}
\centering
\begin{tabular}{c | c c c c | c c c c}
  & \multicolumn{4}{c|}{\textbf{Single RL}}  & \multicolumn{4}{c}{\textbf{Grid-switching RL}} \\
  Reference force (x, y) [N] & x error & x std & y error & y std  & x error & x std & y error & y std \\
 \hline
 (1, -1) & 1.07 & 0.458 & 0.871 & 0.317 & 0.532 & 0.520 & 0.382 & 0.362 \\
 (2, -1) & 0.953 & 0.436 & 0.351 & 0.274 & 0.298 & 0.218 & 0.935 & 0.207 \\
 (2, 0)  & 0.419 & 0.281 & 0.369 & 0.269 & 0.307 & 0.274 & 0.520 & 0.463 \\
 (3, 0)  & 0.394 & 0.338 & 1.12 & 0.386 & 0.467 & 0.402 & 0.830 & 0.331 \\
 (1, 1)  & 1.11 & 0.515 & 0.690 & 0.381 & 0.440 & 0.220 & 0.302 & 0.214 \\
 (2, 1)  & 1.02 & 0.421 & 0.714 & 0.727 & 0.696 & 0.369 & 0.269 & 0.182 \\
 \hline
 \textbf{Overall mean} & \textbf{0.828} & \textbf{0.408} & \textbf{0.685} & \textbf{0.392} & \textbf{0.435} & \textbf{0.325} & \textbf{0.537} & \textbf{0.201} 
\end{tabular}
\end{minipage}
\end{table*}

\begin{figure*}[t]
\includegraphics[clip, width=18cm]{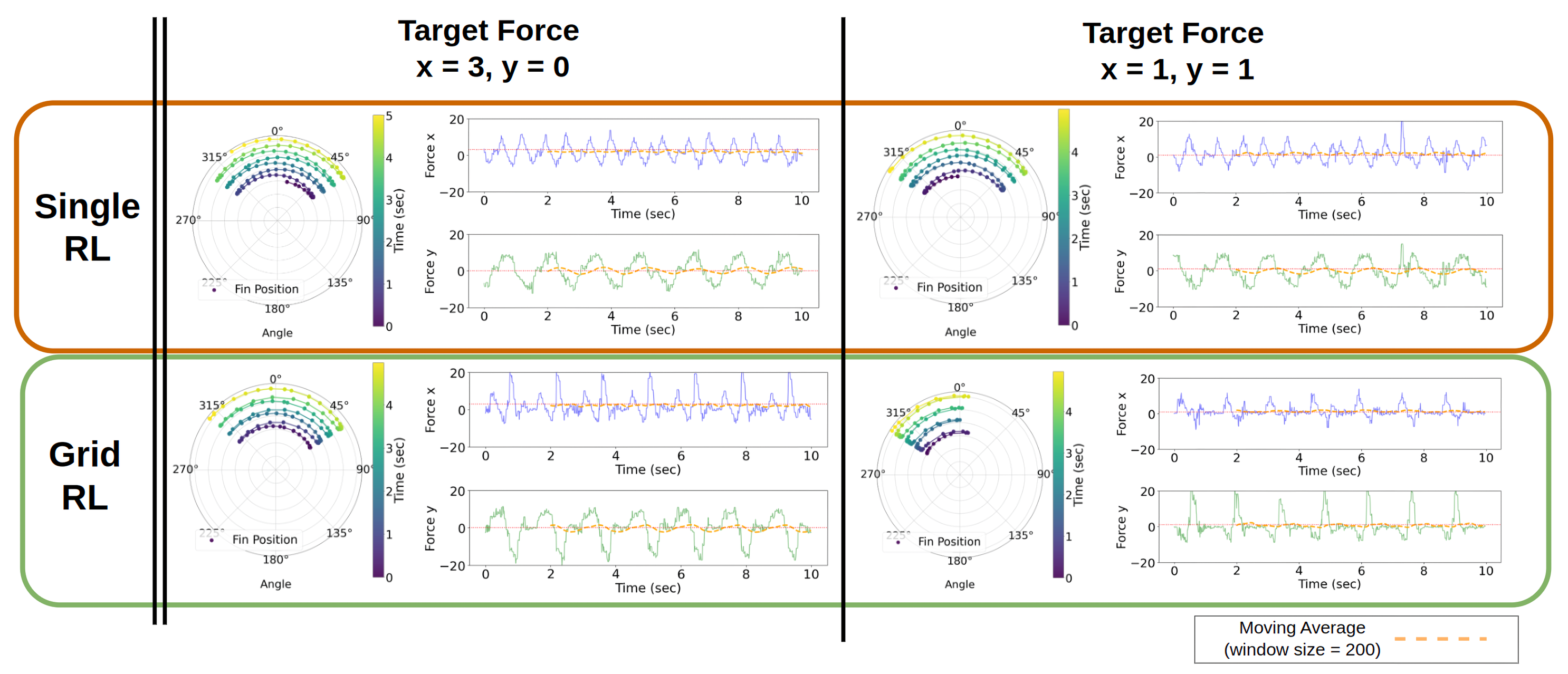}
\caption{Evaluation results. In each test, the figure on the left shows the movement trajectory of the fin in polar coordinates (degrees). Data from the first 5 seconds of the experiment were used as representative. The right-hand figure shows the actual force output in each axis (x, y). This was recorded by the force sensor and the first 10 seconds were plotted as representative. The dashed line shows the moving average of the forces (window size 200).}
\label{eval}
\end{figure*}

\subsubsection{Motion controllers}

To achieve the desired force generation, we propose two different motion control frameworks. Figure \ref{fig:RL} shows the fundamental structures of each design. The first approach involves training a single RL controller using a randomly defined reference force in the $x$ and $y$ directions. In this framework, the RL agent generates motion based on the current state, which includes the reference forces. This approach allows the RL agent to dynamically adapt its motion strategy based on the changing states, enabling generalized force control across a wide range of conditions. A key advantage of this method is that it requires less training effort, as only one controller is trained to handle multiple force references. Training for this model was carried out in 30,000 steps.

The second approach is the grid-switching motion controller. In this framework, multiple models are independently trained for different ranges of reference forces. During operation, the appropriate model is selected based on the current reference force. Although this approach requires more training time compared to a single controller, it offers higher precision since each model is specialized for a specific range of forces. The use of a surrogate simulator as the training environment makes this method feasible, as it significantly reduces the time and resources needed for model training by simulating the dynamics of the system without requiring physical experiments. The structure of the neural network in the RL is the same as in the single general RL model. Each model was trained in 10,000 steps.

\section{Results and discussion}
\subsection{Surrogate model performance}
The forward models inference was calculated on three different test datasets with different experimental conditions. The chosen metrics for evaluation are Root Mean Squared Error (RMSE), Mean Absolute Error (MAE) and Dynamic Time Warping (DTW) \cite{dtwintro}, \cite{dtw2}. The average result of this evaluation is shown in Table \ref{tab:inf}. DTW calculates the optimal alignment between two time-series by stretching or compressing parts of the series to minimize the distance between them. This flexibility makes DTW especially useful in cases where patterns in the data may have temporal shifts but still capture the general characteristic as time-series continuous data. By using DTW, we can capture the similarity between time-series that may not be perfectly synchronized but exhibit similar temporal behaviors. A DTW value close to zero means that the model successfully captures the temporal dynamics of the system, with minimal distortion required to align the predicted and actual sequences. This low DTW value suggests that the model performs well not only in predicting individual data points as evaluated by RMSE and MAE but also in capturing the underlying time-series behavior over the entire sequence, including any shifts or delays. 

\subsection{Performance of the motion in real-world}
The controller trained in the surrogate simulator was tested with the real fin. The equipment for the experiment was the same as the data collection as shown in Fig. \ref{fig:fin_axes}. The actual force generated was also collected with a force sensor in the same way as the data collection setups for performance evaluation. The experiments were carried out on six different reference forces, each with the two types of RL motion controller described in the previous section. In each experiment, 30 seconds of data from the beginning of the experiment were used as the valid interval for evaluation. The data were recorded at 100 Hz. The conditions of the experiment and the results are summarized in Table \ref{tab:eval}. As shown in Table \ref{tab:eval}, the grid RL motion controller achieved a smaller error than a single RL. Figure \ref{eval} shows the reference results. The figures on the left in each section show the fin trajectory and the right shows the measured force in each direction. As shown in Fig. \ref{eval}, the grid RL design achieves a more optimal motion that moves in the opposite direction to the reference force (x=1, y=1). 

\subsection{Discussion}
As shown in Table. \ref{tab:eval} and Fig. \ref{eval}, although the errors in the force output are smaller in the moving average, the variation in the instantaneous force output at each instant is still large. This is due to the specification of the fin actuator. It is physically difficult to continuously output a uniform force, as in the case of a propeller. This is because the vector of forces that a fin can output in a single action depends on the latest position of the fin. It is difficult to output forces only in the desired direction in a single action. Therefore, the motion trajectory in Fig. \ref{eval} shows that the RL agent has learned to move to produce a force that cancels the output in the undesired direction to obtain a reference force in the window under these conditions.
As a limitation, the data size deployed in memory by one model per grid-switching design was approximately 7.2 MB with our setups. There is a limitation that the memory requirement increases with increasing grid dimension and model complexity.

\section{Conclusion}
This study presents a new framework for motion control in soft fin actuators. Our method combines a DNN based surrogate model with a RL agent. In this framework, the surrogate model serves as a simulator, accurately estimating the complex relationship between fin motion and the generated force It allows the RL agent to train effectively in a simulated environment. This approach reduces the need for physical tests in environments such as underwater experiments, making learning faster and simpler. The design of the reward function minimizes the error between the generated and target forces and incorporates the Sobolev norm to ensure smooth force output. Additionally, grid-switching motion control enables optimized RL models for specific force reference ranges, achieving highly optimal motion. This grid-switching approach requires a large amount of training. Therefore, it is more compatible with surrogate-based simulation, where the training setup is easier and fast-forwardable than training in a real environment.

In evaluation experiments, the RL agent trained within the surrogate simulation was applied to the actual fin actuator in the real world to evaluate the control accuracy in different force reference settings. The grid-switching model consistently demonstrated small error margins in comparison to a single RL controller. The results show that the use of a simulator with a surrogate model can produce the forces using the soft fin actuator required for robot movement in the real world. Future work will integrate our method with body-frame force control for real fin-actuated robots which have multiple soft fins.

\section*{Appendix}
The scripts and dataset are available here: \url{https://github.com/Centre-for-Biorobotics/ML-soft-fin-motion-RoboSoft2025}

%\vfill

% Can be used to pull up biographies so that the bottom of the last one
% is flush with the other column.
%\enlargethispage{-5in}

\bibliographystyle{splncs04}
% Generated by IEEEtran.bst, version: 1.14 (2015/08/26)

% that's all folks
\end{document}